\def\year{2022}\relax
\documentclass[letterpaper]{article} 
\usepackage{aaai22}  
\usepackage{times}  
\usepackage{helvet}  
\usepackage{courier}  
\usepackage[hyphens]{url}  
\usepackage{graphicx} 
\usepackage{natbib}  
\usepackage{caption} 
\DeclareCaptionStyle{ruled}{labelfont=normalfont,labelsep=colon,strut=off} 
\usepackage[switch]{lineno} 
\usepackage{enumerate}
\usepackage{makecell}
\usepackage{multirow}
\usepackage{booktabs,tabularx}
\usepackage{amsmath,amssymb}
\usepackage{algorithm}
\usepackage{algpseudocode}

\usepackage{newfloat}
\usepackage{listings}
\floatstyle{ruled}
\newfloat{listing}{tb}{lst}{}
\floatname{listing}{Listing}
\title{FedOCR: Efficient and Secure Federated Learning for Scene Text Recognition}
\author{
    Wenqing Zhang\textsuperscript{\rm 1},
    Yang Qiu\textsuperscript{\rm 1},
    Song Bai\textsuperscript{\rm 1},
    Rui Zhang\textsuperscript{\rm 2},
    Xiaolin Wei\textsuperscript{\rm 2},
    Xiang Bai\textsuperscript{\rm 1}
}
\affiliations{

    \textsuperscript{\rm 1}Huazhong University of Science and Technology,
    \textsuperscript{\rm 2}Meituan

    \{wenqingzhang, yqiu, xbai\}@hust.edu.cn, songbai.site@gmail.com, \{zhangrui36, weixiaolin02\}@meituan.com
}

\usepackage{bibentry}

\begin{document}

\maketitle

\begin{abstract}
While scene text recognition techniques have been widely used in commercial applications, data privacy has rarely been taken into account by this research community. Most existing algorithms have assumed a set of shared or centralized training data. However, in practice, data may be distributed on different local devices that can not be centralized to share due to privacy restrictions. 
In this paper, we study how to make use of decentralized datasets for training a robust scene text recognizer while keeping them stay on local devices. 
To the best of our knowledge, we propose the first framework leveraging federated learning for scene text recognition, which is trained with decentralized datasets collaboratively. Hence we name it FedOCR. To make FedCOR fairly suitable to be deployed on end devices, we make two improvements including using lightweight models and hashing techniques. We argue that both are crucial for FedOCR in terms of communication efficiency and security for federated learning.
The simulations on decentralized datasets show that the proposed FedOCR achieves competitive results to the models that are trained with centralized data, with fewer communication costs and higher-level privacy-preserving.
\end{abstract}

\section{Introduction}
Text in scene images contains valuable semantic information for text reading and has become one of the most popular research topics in academia and industry for a long time~\cite{goel2013whole,almazan2014word,su2014accurate,luo2019moran,li2019show,zhangautostr,yu2020towards}. 
In practice, scene text recognition has been applied to various real-world scenarios, such as autonomous navigation, photo transcription, and scene understanding. 
With the development of deep learning and the emergence of public text datasets, significant progress on scene text recognition has been made in recent years.


\begin{figure*}
\begin{center}
\includegraphics[width=12cm] {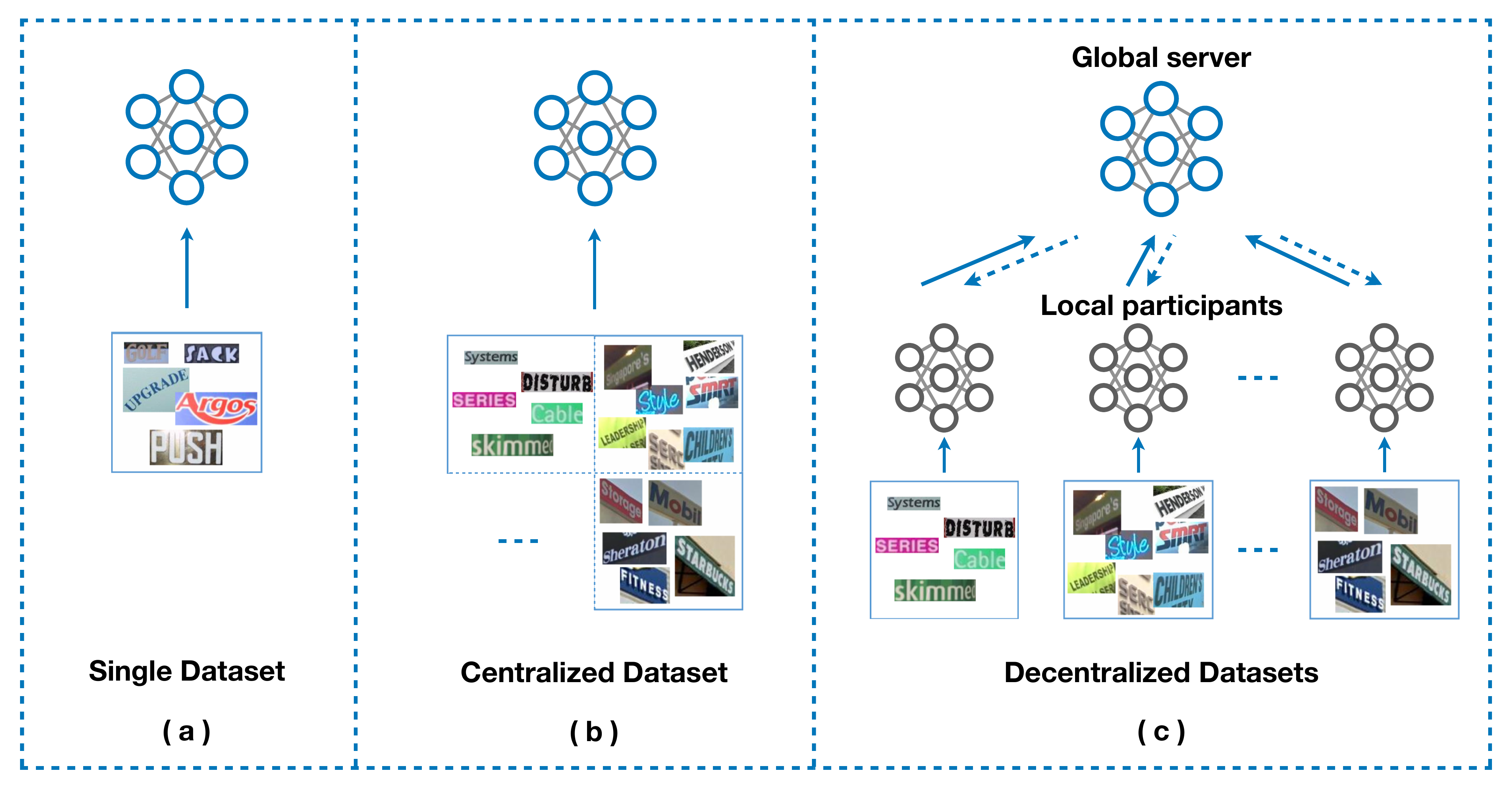}
\end{center}
  \caption{An illustration of training scene text recognizers with (a) a single dataset, (b) a centralized dataset from different devices, and (c) decentralized datasets distributed on different local devices.}
\label{fig-inro}
\end{figure*}

However, most of the existing scene text recognition algorithms assume that a large scale set of training images is easily accessible. As shown in Fig.~\ref{fig-inro}(a), in real conditions, algorithms may achieve sub-optimal performance and be unable to model the data variations or diversity owing to the lack of sufficient images. To remedy this, some works~\cite{bartz2019kiss,hu2020gtc} merge different public datasets to build a more robust text recognizer, as illustrated in Fig.~\ref{fig-inro}(b). However, centralizing data in this way is simply problematic in~\textbf{many real-world scenarios}. For example, many laws and regulations strengthening the data privacy constrain the use of data stored on local devices, such as General Data Protection Regulation (GDPR)~\cite{voigt2017eu}. Besides, centralizing tremendous image data from different local devices incurs heavy communication loads. That means it is simply intractable to centralize large amounts of data for scene text recognition training in practice. Our solution, which works within the framework of federated learning, is illustrated in Fig.~\ref{fig-inro}(c).

Federated Learning (FL), a new concept first proposed by McMahan \emph{et al.}~\cite{mcmahan2016communication}, allows data owners to train a shared model collaboratively while keeping data stored on different local devices. However, directly applying FL to scene text recognition faces two inevitable difficulties. 
First, in most scene text recognition algorithms, a heavyweight backbone model is usually adopted for the sake of better performance. Hence, it results in heavy burdens of the parameter transmission while doing federated learning. Second, there is an extra computational cost from a privacy-preserving module to handle privacy leakage due to the honest-but-curious global server in general federated learning frameworks.

\begin{figure*}[t]
\begin{center}
\includegraphics[width=12cm] {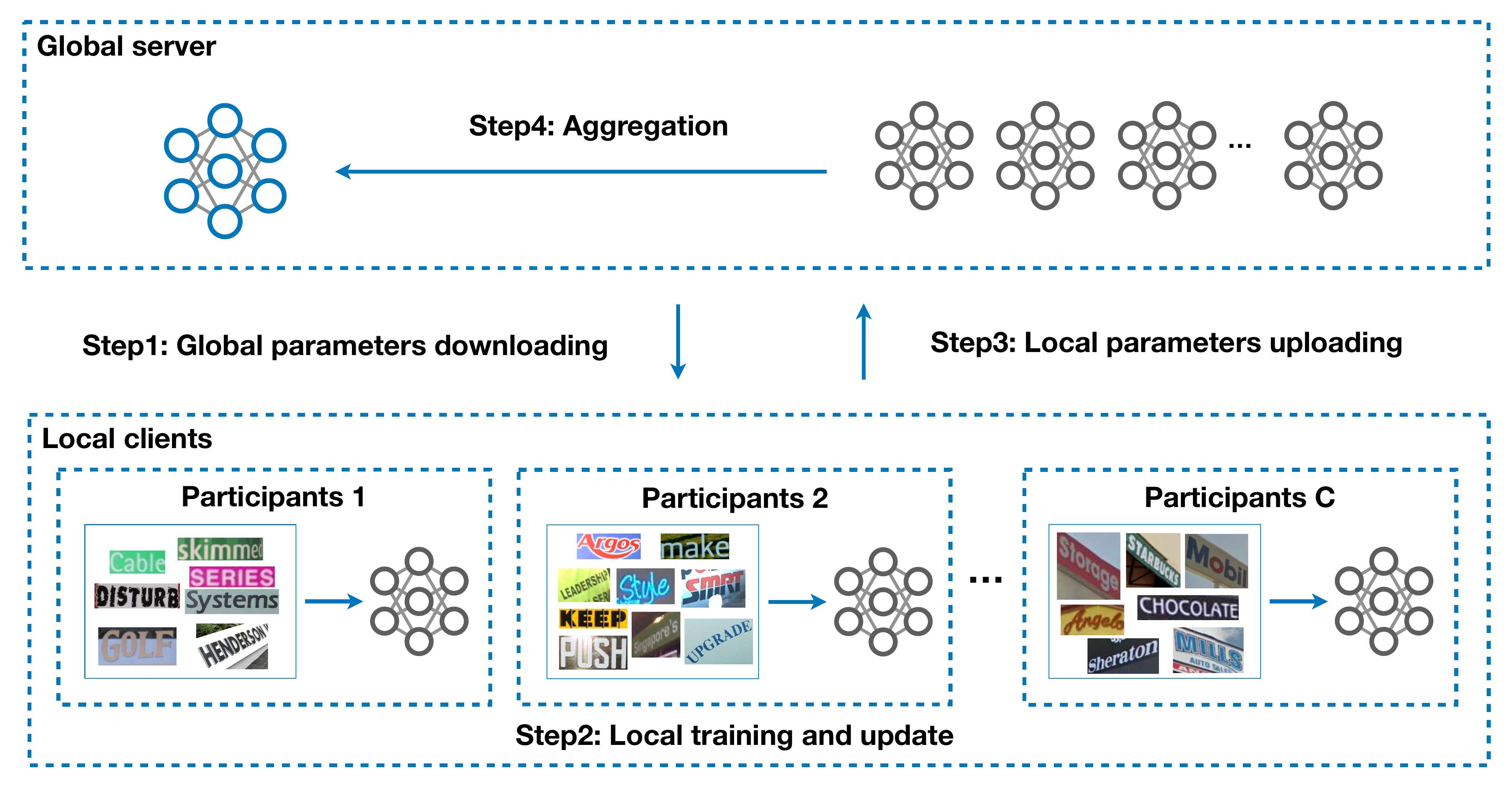}
\end{center}
  \caption{The pipeline of our federated scene text recognition framework.}
\label{fig-structure}
\end{figure*}

In this paper, to the best of our knowledge, we propose the first federated learning framework for scene text recognition, which we name FedOCR.
In our FedOCR (a schematic is given in Fig.~\ref{fig-structure}), all participants train a shared model collaboratively without centralizing the training images.
In this manner, datasets on different local devices have an indirect influence on the training of the global model, which leads to a competitive performance to the model trained with a centralized set of data. To improve the communication efficiency between the global server and local clients, we argue two important aspects in FedOCR, ~\emph{i.e.}, lightweight models and hashing techniques. Moreover, benefited from the hashing technique, we can avoid privacy leakage to the global server by a specific hashing function and the random seeds, which saves an extra computational cost for a privacy-preserving module. As a consequence, the proposed FedOCR is readily to be deployed in practical applications for scene text recognition.

Compared with existing scene text recognition methods~\cite{luo2019moran,li2019show,bartz2019kiss,zhan2019esir,yue2020robustscanner,bhunia2021joint} without federated learning, the proposed framework has the following intriguing merits. 
First, FedOCR can make use of more abundant image data from different local devices. 
Particularly, there are billions of end devices with tremendous text images benefiting scene text recognition. Therefore, our framework may have great potential in real-world applications of scene text reading.
Second, by design, our framework has a superior trade-off between parameter transmission efficiency and performance. The proposed text recognizer has much fewer parameters than existing scene text recognition algorithms but encouragingly reaches a comparable performance. 
Last, it can encrypt and decrypt with the hashing technique, which provides higher-level privacy-preserving without an extra computational cost.

In summary, the main contributions of this paper are three-fold.
\begin{itemize}
\item We reveal the problem of data privacy in scene text recognition, which is somehow overlooked by the existing methods.
\item We propose the first federated scene text recognition framework called FedOCR for training a recognizer with decentralized datasets distributed on different local devices.

\item FedOCR is a highly communication-efficient as well as privacy-preserving framework by incorporating lightweight backbones and hashing techniques, which makes it suitable to be deployed in real privacy-sensitive applications and edge devices.
\end{itemize}

\section{Related Work}
Scene text recognition has attracted great interest for a long time. According to Long \emph{et al.}~\cite{long2018scene}, representative methods can be roughly divided into two mainstreams, ~\emph{i.e.}, Connectionist Temporal Classification (CTC) based and attention-based methods. Generally, the CTC-based methods model scene text recognition as a sequence recognition task. For example, Shi \emph{et al.}~\cite{shi2016end} combine the convolutional neural network (CNN) with the recurrent neural network (RNN) to extract sequence features from input images, and decode the features with a CTC layer. Different from Shi \emph{et al.}~\cite{shi2016end}, Gao \emph{et al.}~\cite{gao2019reading} use stacked convolutional layers to extract contextual information from inputs without RNN, and show advantages with low computational costs. Zhang \emph{et al.}~\cite{zhang2020adaptive} turn
text recognition into a visual matching problem by exploiting the repetition of glyphs in language, and build this similarity between units into the proposed architecture. 
Meanwhile, attention-based methods extract features more effectively via the attention mechanism. For instance, Liu \emph{et al.}~\cite{liu2018squeezedtext} propose a binary convolutional encoder-decoder network to provide real-time scene text recognition. Liu \emph{et al.}~\cite{liu2018char} propose a character-aware neural network with a hierarchical attention mechanism, which adopts a local transformation to rectify characters individually. Unlike other attention-based algorithms, Bai \emph{et al.}~\cite{bai2018edit} propose Edit Probability (EP) to handle the misalignment between the output sequence of probability distribution and the ground-truth sequence. Nguyen \emph{et al.}~\cite{nguyen2021dictionary} incorporate a dictionary in both the training and inference stage to make a more robust scene text recognition system.


Undoubtedly, large amounts of real-world data are needed in practical applications of those scene text recognition methods. However, tremendous image datasets are distributed on different companies, communities or local devices, and can not be centralized to share. To handle this problem, McMahan \emph{et al.}~\cite{mcmahan2016communication} first propose the concept of Federated Learning (FL) to train deep networks from decentralized data collaboratively. 
Following McMahan \emph{et al.}~\cite{mcmahan2016communication}, many researchers are working on improving the federated learning with more efficient parameter transmission and higher-level privacy-preserving. 
To improve privacy security, Wei \emph{et al.}~\cite{wei2019federated} propose a federated learning framework based on differential privacy, in which artificial noises are added to the local parameters of participants before the model aggregation. Sun \emph{et al.}~\cite{sun2021soteria} develop an effective defense called Soteria by perturbing data representations to against model inversion attack in FL.
To improve communication efficiency, Reisizadeh \emph{et al.}~\cite{reisizadeh2019fedpaq} propose a communication-efficient federated learning method with periodic averaging and quantization. Gao \emph{et al.}~\cite{gao2021convergence} propose the error-compensated double compression mechanism to significantly reduce the communication cost. \textbf{Especially very recently, the computer vision community starts to pay attention to federated learning, thus arising several pioneering works.} For example, Luo \emph{et al.}~\cite{luo2019real} implement object detection algorithms with federated learning and release a reliable benchmark framework. In the medical field, Zhu \emph{et al.}~\cite{zhu2019privacy} implement a privacy-preserving federated learning system with the differential privacy for brain tumor segmentation. 
Li \emph{et al.}~\cite{li2021model} conduct contrastive learning in model-level to correct the local training of individual parties to achieve high performance in image classification tasks.
To the best of our knowledge, we propose the first federated scene text recognition framework, which is more efficient in communication and provides higher-level privacy-preserving.

\section{Methodology}
In this section, we first introduce the pipeline of our federated scene text recognition framework. Then, we describe the details of local training and global aggregation, which are the two main steps in federated learning. Finally, we elaborate on how to improve communication efficiency and preserve data privacy in our framework.

\subsection{Pipeline of FedOCR}

According to Yang \emph{et al.}~\cite{yang2019federated}, our framework is a kind of horizontal federated learning, where datasets of different participants share the same feature space but differ in samples. 
Suppose we have $C$ data owners, which have different sets of training images $\{D_1,...,D_C\}$. We denote the accuracy of the text recognizer trained with decentralized datasets $\{D_1,...,D_C\}$ as $Acc_{FED}$. Note that these decentralized datasets are not shared or transferred to other participants during training.
We denote the accuracy of the text recognizer trained with a centralized dataset $D=D_1 \cup ... \cup D_C$ as $Acc_{SUM}$. Basically, the objective of FedOCR is to minimize the difference between $Acc_{FED}$ and $Acc_{SUM}$. A smaller difference between $Acc_{FED}$ and $Acc_{SUM}$ means a better performance of our federated learning for scene text recognition.

Fig.~\ref{fig-structure} illustrates the pipeline of our federated scene text recognition framework. There are $C$ participants, each of which has a set of data containing cropped text word images and transcriptions, and a global server for local model parameter aggregation. We assume all participants agree in advance on the same network architecture and the same training objective but do not share their datasets.
The whole learning process can be decomposed into four steps: 
\begin{enumerate}
    \item[(1)] Before each round of local training, all participants start with the same parameters, which are initialized randomly in the first round and downloaded from the global server in the next rounds.
    \item[(2)] Each participant trains the model with its dataset for $E_l$ epochs individually.
    \item[(3)] All participants calculate parameter increments compared to the original parameters in a round, and all parameter increments are sent to the global server.
    \item[(4)] The global server aggregates all parameter increments by average, and updates a set of global parameters. Before the next local training, the global parameters are downloaded for local model updating.
\end{enumerate}
Following this pipeline, our federated training continues until convergence.

\subsubsection{Local Training.} In our FedOCR, each participant $i$ and the global server maintain a set of local model parameters $W^i$ and $W^{global}$, respectively. Algorithm~\ref{code:local-training} describes the local training process of our framework. As shown, all participants first download the latest global parameters from the global server and overwrite their local parameters. 
Then, participants train local models with their datasets independently for $E_l$ epochs and send parameter increments to the global server. During local training, all participants do not share any image data with others. 
To update the global parameters efficiently, all participants should train their models enough before parameter transmission. McMahan \emph{et al.}~\cite{mcmahan2016communication} demonstrate that sufficient epochs of local training can bring a dramatic increase in parameter update efficiency. Detailed experiment settings of our FedOCR are provided in the next section.

\begin{algorithm}[!tb]
  \caption{Local Training}
  \begin{algorithmic}[1]
    \Require
        Latest global parameters $W_{t}^{global}$ in round $t$;
        Local training learning rate $\eta^i, i\in [0,C-1]$
    \For{each $i\in [0,C-1]$}
        \State Overwrite local weight vectors: $W_{t}^i = W_{t}^{global}$
    \EndFor
    \ForAll {local participant $i\in \{0,1,...,C-1\})$}
      \For{$e \in [0, E_l-1]$}
        \For{$s \in [0, step_{max}]$}
            \State Sample a minibatch $B_{s}$
            \State Compute gradients: $g_t^i = \bigtriangledown{L(B_{s};W_{t}^i)}$
            \State Update local parameters: $W_t^i = W_t^i - \eta^i \cdot g_t^i$
        \EndFor
      \EndFor
      \State Compute local parameter increments: 
      \State $\Delta{W_t^i} = W_t^i - W_{t}^{global}$
      \State Send $\Delta{W_t^i}$ and data size $S_i$ to the global server
    \EndFor
  \end{algorithmic}
  \label{code:local-training}
\end{algorithm}

\subsubsection{Global Aggregation.} To aggregate parameter increments from different local participants, McMahan \emph{et al.}~\cite{mcmahan2016communication} propose a straightforward approach to aggregate all local participants' parameters by average. Following steps in Algorithm~\ref{code:global-server}, we adapt the federated average method~\cite{mcmahan2016communication} to our federated scene text recognition framework. In the global aggregation step of our FedOCR, we average all parameter increments and update former global parameters, which are available for all participants' downloading.

\begin{algorithm}[!tb]
  \caption{Global Aggregation}
  \begin{algorithmic}[1]
    \Require  
        All local parameter increments $\{ \Delta{W_t^i} | i\in [0,C-1]\}$ in round $t$;
        Local data size $\{ S_i | i\in [0,C-1]\}$
        Global parameters $W_{t}^{global}$;
    \State Compute global parameter increments: 
    
     $\Delta{W_t^{global}} = ( ~\sum_{i=0}^{C-1} S_i \Delta{W_t^i} ~) ~/~ \sum_{i=0}^{C-1} S_i$
    \State Update global parameters: 
    
    $W_{t+1}^{global} = W_{t}^{global} + \Delta{W_t^{global}}$
    \State Send $W_{t+1}^{global}$ to all participants
  \end{algorithmic}
  \label{code:global-server}
\end{algorithm}

\subsection{Communication Efficiency}

Communication efficiency is an essential property in federated learning. 
For instance, if the size of one participant's model is one hundred megabytes, tens of gigabytes will be required to transmit in a round, when hundreds of clients participate in a federated learning framework.
Under such a circumstance, plenty of parameters result in huge communication costs, which lead to a training bottleneck. 
To reduce communication burdens, we replace the heavyweight backbone, such as ResNet~\cite{he2016deep}, for feature extraction in text recognizers with a lightweight neural network. To further decrease the parameter size, we extend a hashing technique~\cite{chen2015compressing} to compress the parameters of CNN and RNN, which makes it applicable for any text recognizer. 
In this way, the text recognizer in our FedOCR has much fewer parameters compared with existing text recognition algorithms, which shows great potential in practical federated learning deployment.

\subsubsection{Hashing Technique.} In fact, any well-designed scene text recognition model can be applied in our federated learning framework. However, considering the communication efficiency, the network with fewer parameters is more appropriate and practical. Therefore, we propose to compress model parameters by a hashing technique.
Specifically, we compress network parameters in a weight sharing manner that a random subset of parameters in a layer share the same parameter. Following Algorithm~\ref{code:compression}, we can compress the parameters in a scene text recognition network with a hyper-parameter $\gamma$ to control the compression ratio, and it can reduce the parameter size to a large extent. It should be noted that $\lfloor e \cdot \gamma \rfloor$ means the largest integer that is smaller than $e \cdot \gamma$ in Algorithm~\ref{code:compression}. 
Notably, the specific hashing function and the random seeds are shared among all local participants to keep the same relationship between real weight vectors and virtual weight matrices of all local models.

\begin{algorithm}[!tb]
  \caption{Hashing Technique}
  \begin{algorithmic}[1]
    \Require  
        Compression ratio $\gamma$; 
        Hashing seeds \{$seed^l | l \in [0,L-1]$\}, where $L$ is the number of network layers;
    \Ensure
        A compressed network;
    \For{each layer $l$ in the entire network}
        \State Assume the total size of weight matrix $W^l$ is $T^l$
        \State Generate a real weight vector $R^l$ with a size $T^l * \gamma$
        \State Generate a random sort $RS^l$ of numbers from $0$ to $T^l - 1$ with a hashing function and a seed $seed^l$
        \State Generate an index vector $I^l$: $[\lfloor e \cdot \gamma \rfloor, \text{for e in } RS^l]$
        \State Reshape $I^l$ as the shape of $W^l$
        \State Generate a virtual weight matrix: $V^l = R^l[I^l]$
    \EndFor
    \State Initialize our network with $R^l, l \in [0,L-1]$, and the total parameter size is compressed to $\gamma \cdot \sum_{l=0}^{L-1} T^l$
  \end{algorithmic}
  \label{code:compression}
\end{algorithm}

\subsubsection{Text Recognizer.} Following the above methods, we can improve any existing text recognition algorithms to construct a lightweight text recognizer. 
Specifically, in our experiments, we optimize a classical text recognizer, ASTER~\cite{shi2018aster}. We replace the encoder in ASTER with ShuffleNetV2~\cite{ma2018shufflenet} and apply the hashing technique to the entire model parameters. 
Benefited from hashing techniques and lightweight networks, we successfully decrease communication costs to a large extent in our federated learning framework.

Moreover, we keep the network structure and experiment settings the same with ASTER as much as possible.
Similar to ASTER, the text recognizer in our experiments consists of a rectification network, a lightweight convolutional encoder, and an attentional sequence-to-sequence model. 
We briefly introduce the method of scene text recognition as follows:
Firstly, an input image is rectified by a rectification network before being sent into a recognition network. The rectification network based on the Spatial Transformer Network (STN) aims to rectify perspective or curved texts.
Secondly,  we use a lightweight neural network as the encoder to extract the feature sequence from the rectified image.
Lastly, we use an attentional sequence-to-sequence model as the decoder to translate the feature sequence. 
During inference, we use beam searching by holding five candidates with the highest accumulative scores at every step.

\subsubsection{Network Training.} After neural network initialization, the mapping relationship between real weight vectors and virtual weight matrices is fixed, which is defined in Algorithm~\ref{code:compression}. In the forward computation, it is the virtual weight matrices that participate in calculation with input features. In the backward propagation, the gradients of all parameters in real weight vectors are calculated based on virtual weight matrices' parameter gradients.

Let $V_{i,j}^l$ denote the $i$-th row and $j$-th column element of a virtual weight matrix at layer $l$, and let $R_k^l$ denote the $k$-th element in the corresponding real weight vector. Assuming that

\begin{equation}
    \frac{\partial{\mathcal{L}}}{\partial{V_{i,j}^l}} = g_{i,j}^l\text{, }
\end{equation}
where $g_{i,j}^l$ is computed from the loss. Moreover,
\begin{equation}
    \frac{\partial{V_{i,j}^l}}{\partial{R_k^l}} = \mathbb{I}(I^l[i,j], k)
    \text{, and }\mathbb{I}(a,b) = 
    \begin{cases}
        1& \text{if } a=b, \\
        0& \text{otherwise}
    \end{cases}
\end{equation}
Based on the above equations, we can obtain any parameter's gradient in the real weight vector as follows:

\begin{equation}
    \begin{aligned}
        \frac{\partial{\mathcal{L}}}{\partial{R_k^l}} &= \sum_{i}\sum_{j}\frac{\partial{L}}{\partial{V_{i,j}^l}} \cdot \frac{\partial{V_{i,j}^l}}{\partial{R_k^l}} \\
        &= \sum_{i}\sum_{j}g_{i,j}^l \cdot \mathbb{I}(I^l[i,j], k)\text{. }
    \end{aligned}
\end{equation}

\subsection{Privacy Preserving}
Federated learning can provide training procedures at a high level of security, but the global server still has a chance to compromise data privacy, such as model inversion~\cite{fredrikson2015model} and GAN-based attacks~\cite{hitaj2017deep}. 
Usually, local network parameters or their increments are sent to the global server in each communication round, which gives the honest-but-curious server a chance to spy on local sets of data.
In recent works, Geiping~\emph{et al.}~\cite{geiping2020inverting} and Phong \emph{et al.}~\cite{phong2018privacy} show that the gradients may reveal information of training samples and apply an additively homomorphic encryption scheme to their federated framework.
Shokri \emph{et al.}~\cite{shokri2015privacy} propose to upload partially gradients added with noise to avoid information leakage, and apply differential privacy to parameter updates for a higher level of security. 
However, the above methods bring more computational costs or a dramatic decrease in accuracy because of the privacy-preserving module.

In our FedOCR, we adopt the hashing technique to compress the entire model parameters with a hashing function and random seeds.
They are equivalent to an encryption-decryption module and the keys, but it can save much computational costs without the encryption-decryption step.
For the parameter aggregation in the global server, we only upload increments of the parameters in real weight vectors, which can not be used to reconstruct the complete network without the specific hashing function and the random seeds. 
As for all local participants, they share the same hashing function and random seeds, so the average operation in the global aggregation can be directly applied to these parameter increments. 
Therefore, the global server can not compromise the private data, while it can finish its global aggregation task.
Moreover, the other attackers also can not do anything with the compressed parameter increments, because they mean nothing without hashing functions and seeds.
In this way, we enhance the privacy-preserving in our FedOCR without introducing an extra computational cost.

\section{Experiments}

\subsection{Datasets}
Two synthetic datasets~\cite{jaderberg2014synthetic,gupta2016synthetic} and six public real-world datasets are used to train local models, and our models are evaluated on seven general datasets. In our federated settings, we construct different local datasets with the public real-world datasets. These datasets are briefly introduced as follows:

\vspace{1ex}\noindent\textbf{Synth90k}~\cite{jaderberg2014synthetic} contains 9 million images generated from a set of 90k English words. Words are rendered onto natural images with random transformations and effects.

\vspace{1ex}\noindent\textbf{SynthText}~\cite{gupta2016synthetic} contains 0.8 million images for end-to-end text detection and recognition tasks. Therefore, we crop word images using the ground-truth word bounding boxes.

\vspace{1ex}\noindent\textbf{ICDAR 2003 (IC03)}~\cite{lucas2005icdar} contains 860 cropped word images for evaluation after discarding images that contain non-alphanumeric characters or have fewer than three characters, which follows~\cite{mishra2012top}. For training, we use 1150 cropped images after filtering.

\vspace{1ex}\noindent\textbf{ICDAR 2013 (IC13)}~\cite{karatzas2013icdar}, which inherits most images from IC03 and extends it with new images, contains 1015 cropped word images for evaluation after filtering. For training, we use 848 cropped images after filtering.

\vspace{1ex}\noindent\textbf{ICDAR 2015 (IC15)}~\cite{karatzas2015icdar} contains images captured by a pair of Google Glasses casually, and many images are severely distorted or blurred. For a fair comparison, we evaluate models on 1811 cropped word images after filtering. For training, we use 4426 cropped images after filtering.

\vspace{1ex}\noindent\textbf{IIIT5K-Words (IIIT5K)}~\cite{mishra2012top} contains 3000 word images collected for evaluation and 2000 word images for training, which are mostly horizontal text images.

\vspace{1ex}\noindent\textbf{Street View Text (SVT)}~\cite{wang2011end} is collected from the Google Street View, and it contains 647 images of cropped words, many of which are severely corrupted by noise, blur, or low resolution.

\vspace{1ex}\noindent\textbf{Street View Text Perspective (SVTP)}~\cite{quy2013recognizing}, which is collected from Google StreetView and contains many distorted images, contains 645 word images for evaluation.

\vspace{1ex}\noindent\textbf{CUTE80 (CUTE)}~\cite{risnumawan2014robust} contains 80 real-world curved text images with high quality. For evaluation, we crop 288 word images according to its ground-truth.

\vspace{1ex}\noindent\textbf{ArT}~\cite{chng2019icdar2019} is a combination of Total-Text, SCUT-CTW1500, and Baidu Curved Scene Text, which contains images with arbitrary-shaped texts. For training, we use 30271 word images after discarding images that contain non-alphanumeric characters and vertical texts.

\vspace{1ex}\noindent\textbf{COCO-Text}~\cite{veit2016coco} is based on the MS COCO dataset, which contains images of complex everyday scenes. For training, we use 31943 cropped images after discarding images that contain non-alphanumeric characters and vertical texts.

\begin{table*}[tb]
\begin{center}
\resizebox{0.8\textwidth}{!}{
\begin{tabular}{p{3.0cm}p{2.0cm}p{0.8cm}<{\centering}p{3.0cm}<{\centering}p{3.0cm}<{\centering}p{3.0cm}<{\centering}}
\hline
\noalign{\smallskip}
Models & Backbone & $\gamma$ & Param. (M) & Model (MB) & Accuracy (\%)\\
\noalign{\smallskip}
\hline
\noalign{\smallskip}
ASTER-FL & ResNet & - & 20.99 & 80.52 & 91.94\\
FedOCR-Hash$_{1}$ & ShuffleNetV2 & - & 13.34 ($\downarrow 36.45\%$) & 51.37 ($\downarrow 36.20\%$) & 89.08 ($\downarrow 3.11\%$)\\
FedOCR-Hash$_{1/2}$ & ShuffleNetV2 & 1/2 & 6.70 ($\downarrow 68.08\%$) & 26.05 ($\downarrow 67.65\%$) & 86.65 ($\downarrow 5.75\%$)\\
FedOCR-Hash$_{1/4}$ & ShuffleNetV2 & 1/4 & 3.38 ($\downarrow 83.90\%$) & 13.38 ($\downarrow 83.38\%$) & 85.39 ($\downarrow 7.12\%$)\\
FedOCR-Hash$_{1/8}$ & ShuffleNetV2 & 1/8 & 1.72 ($\downarrow 91.81\%$) & 7.05 ($\downarrow 91.24\%$) & 82.58 ($\downarrow 10.18\%$)\\
\hline
\end{tabular}
}
\end{center}
\caption{Parameter size and accuracy comparison between different models in our FedOCR. The accuracy is the average result of all testing datasets. The models size refers to the storage occupied on the hard-disk. $\gamma$ is the compression ratio of the hashing technique, and ``$\gamma=-$" means that we do not apply the hashing technique to the model. The reduction percentages of parameter size, model size, and accuracy compared with ASTER-FL are shown in parentheses respectively.}
\label{table:paramters}
\end{table*}
\begin{table*}[tb]
\begin{center}
\resizebox{0.75\textwidth}{!}{
\begin{tabular}{p{3.0cm}p{2.0cm}p{1.0cm}<{\centering}p{1.0cm}<{\centering}p{1.0cm}<{\centering}p{1.0cm}<{\centering}p{1.0cm}<{\centering}p{1.0cm}<{\centering}p{1.0cm}<{\centering}}
\hline
\noalign{\smallskip}
Models & Training & IIIT5k & SVT & IC03 & IC13 & IC15 & SVTP & CUTE\\
\noalign{\smallskip}
\hline
\noalign{\smallskip}
 & single & 93.7 & 89.0 & 93.7 & 93.8 & 80.6 & 82.3 & 85.4\\
ASTER-FL & centralized & 95.0 & 91.7 & 95.3 & 94.6 & 82.2 & 83.3 & 91.7\\
 & federated & 95.0 & 90.7 & 94.8 & 94.0 & 82.0 & 82.3 & 91.0\\
\hline
 & single & 90.8 & 83.0 & 90.9 & 89.4 & 77.3 & 77.5 & 82.6\\
FedOCR-Hash$_{1}$ & centralized & 93.1 & 86.4 & 92.5 & 92.2 & 79.7 & 80.6 & 86.8\\
 & federated & 92.9 & 86.9 & 92.0 & 91.7 & 79.4 & 80.8 & 86.5\\
\hline
 & single & 89.2 & 83.0 & 90.2 & 88.5 & 75.3 & 73.8 & 77.8\\
FedOCR-Hash$_{1/2}$ & centralized & 91.6 & 83.6 & 91.0 & 90.3 & 77.9 & 75.5 & 82.3\\
 & federated & 91.2 & 84.2 & 91.6 & 90.7 & 77.5 & 76.0 & 82.6\\
\hline
 & single & 87.2 & 79.1 & 87.1 & 86.1 & 73.4 & 71.5 & 77.4\\
FedOCR-Hash$_{1/4}$ & centralized & 89.4 & 81.6 & 89.5 & 88.8 & 75.9 & 74.3 & 81.6\\
 & federated & 89.0 & 81.8 & 89.3 & 89.2 & 76.3 & 75.2 & 81.6\\
\hline
 & single & 83.5 & 74.8 & 84.8 & 81.4 & 70.2 & 71.2 & 73.3\\
FedOCR-Hash$_{1/8}$ & centralized & 86.7 & 78.8 & 86.7 & 86.0 & 72.3 & 71.6 & 79.5\\
 & federated & 86.6 & 80.1 & 87.1 & 85.4 & 72.4 & 71.6 & 79.9\\
\hline
\end{tabular}
}
\end{center}
\caption{Recognition accuracy in different training manners. ``single": The model is trained only with one participant's dataset; ``centralized": The model is trained with a centralized set of image data; ``federated": The global model is trained with decentralized sets of image data in a federated manner. The detailed structures of different FedOCR-Hash are shown in Tab.~\ref{table:paramters}.
}
\label{table:federated}
\end{table*}

\subsection{Experiment settings}
\subsubsection{Decentralized Datasets for Federated Learning} Different local datasets are constructed by public real-world datasets in our experiment settings. We use the training images from IC03~\cite{lucas2005icdar}, IC13~\cite{karatzas2013icdar}, IC15~\cite{karatzas2015icdar}, IIIT5K~\cite{mishra2012top}, ArT~\cite{chng2019icdar2019}, and COCO-Text~\cite{veit2016coco}. 
As a sequence, we have 70638 real-word text images in total. 
To simulate the decentralized datasets distributed on local devices in federated learning, we, as an honest server, should not known the data distribution and whether there is a data bias.
Hence, we randomly split all training images into different sets of image data for $C$ participants,
It should be mentioned that these different sets of image data should not be shared or transferred to other participants during the training procedures.

\subsubsection{Federated Settings.} Some hyper-parameters should be noted in our federated settings: $C$, the number of participants in our federated scene text recognition framework; $\gamma$, the compression ratio of the hashing technique; $E_{l}$, the number of epochs that each local participant trains the model with its dataset before communication with the global server; $B$, the batch size in local training. In our experiments, we set $C = 5$, $E_l = 3$, $B = 512$ and $\gamma \in \{1/2, 1/4, 1/8\}$.

\subsubsection{Baseline and FedOCR-Hash.} In our experiments, we adopt ASTER\footnote{https://github.com/ayumiymk/aster.pytorch}~\cite{shi2018aster} as the text recognition baseline in our FedOCR, which is denoted as ASTER-FL. Then, we replace the encoder in ASTER-FL with ShuffleNetV2~\cite{ma2018shufflenet}, and this variant of ASTER-FL in our FedOCR is denoted as FedOCR-Hash$_1$. To further reduce the parameter size, we apply the hashing technique to compress FedOCR-Hash$_1$ with different ratios $\gamma \in \{1/2, 1/4, 1/8\}$, and these models are denoted as FedOCR-Hash$_\gamma$ in the following paper.

\subsubsection{Implementation Details.} Following the federated settings, we construct $C=5$ participants in our FedOCR for experiments.
In each local training, all models are locally trained via Adadelta~\cite{zeiler2012adadelta} with an initialization learning rate of 1.0, and each participant trains the scene text recognition model with its dataset individually for $E_l=3$ epochs in each round. All word images are trained directly without data argumentation. 
As for the complete federated training process of our FedOCR, each participant trains its model with the two synthetic datasets for 4 rounds, then trains on its real-world dataset for 40 rounds.

The learning rate is decayed to 0.1 and 0.01 at the 5-th round and the 30-th round, respectively.
Following Algorithm~\ref{code:global-server}, in the global aggregation step, the global server aggregates the parameter increments from all participants by average. To simply simulate the communication procedure of federated learning, we replace the parameter transmission between participants and the global server with saving and restoring checkpoints on the hard-disk.

\subsubsection{Evaluation Metric.} In our experiments, we use the case-insensitive word accuracy for evaluation. If the word prediction and the ground-truth are the same in the lower case, the prediction is correct. The recognition accuracy is the percentage of the correct number of total. Furthermore, the objective of FedOCR is to minimize the difference between the accuracy of the text recognizer trained with decentralized datasets and trained with a centralized dataset. A smaller difference means a better performance of our FedOCR.

\subsection{Experiments on FedOCR}
In this subsection, we first compare the parameter reduction and the accuracy decrease of different models in our FedOCR.
Then, we analyze the performance of our FedOCR compared with the other two training manners and show that our FedOCR achieves the objective of federated learning.
Finally, we evaluate the two improvements in communication efficiency of our FedOCR.

\subsubsection{Comparison of Parameter Size and Accuracy.} Tab.~\ref{table:paramters} shows the parameter size and model size of different models in our FedOCR. The accuracy is the average result of all testing datasets. The models size refers to the storage occupied on the hard-disk.
Compared with ASTER-FL, FedOCR-Hash$_{1}$ reduce 36.45\% parameter size, but there is only a 3.11\% accuracy decrease. As for different FedOCR-Hash$_{\gamma}$ in our experiments, FedOCR-Hash$_{1/4}$ with an appropriate compression ratio $\gamma$ achieves a 83.90\% reduction in parameter size and drops only 7.12\% in accuracy. Improved by the lightweight backbone and the hashing technique, the model size of the scene text recognizers in our FedOCR reduces to a large extent, and these lightweight text recognizers encouragingly reach a comparable performance.

\subsubsection{Federated Learning for Scene Text Recognition.} Tab.~\ref{table:federated} shows the detailed results on all testing datasets of ASTER-FL and different FedOCR-Hash$_\gamma$ in three manners of training. 
First, ``single" training means that the model is trained only with one participant's dataset. Second, "centralized" training means that the model is trained with a centralized set of image data. Third, ``federated" training means that the model is trained with decentralized sets of image data in a federated manner. 
As shown in Tab.~\ref{table:federated}, ``federated" and ``centralized" training results of all models are similar to each other and better than ``single" training results. In the ``single" training manner, scene text recognition faces the problem in practice that the image data for training is limited, which causes poor performance in scene text recognition. However, we succeed in training a shared model with decentralized sets of image data collaboratively in the ``federated" training manner, and we do not exchange or expose any image data to other participants. Expectantly, our FedOCR achieves comparable results,
which are very close to the results of the ``centralized" training manner.
Therefore, our FedOCR is effective to train a more robust model without centralizing datasets on different local devices.

\begin{figure}[t]
\begin{center}
\includegraphics[width=1.0\linewidth]{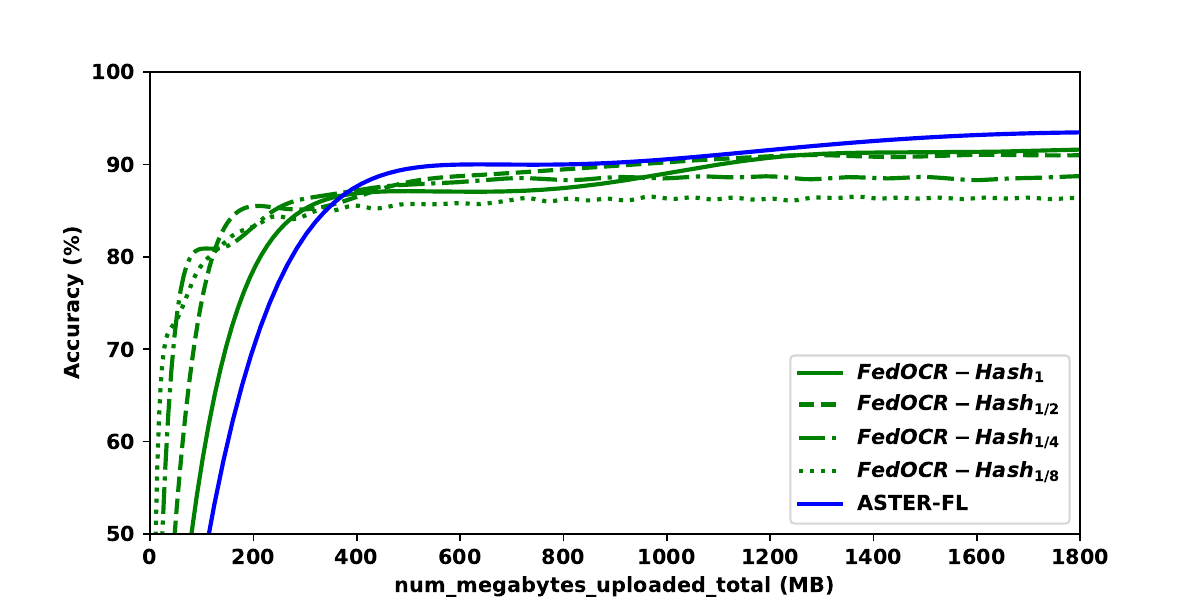}
\end{center}
  \caption{Accuracy on IIIT5k versus number of uploaded megabytes of different models with limited transmitted bytes in federated learning.}
\label{fig:compression}
\end{figure}

\subsubsection{Communication Efficiency Improvement.} In Tab.~\ref{table:federated}, FedOCR-Hash$_{1}$ shows comparable accuracy with ASTER-FL in the ``federated" training manner.
Owing to the lightweight backbone in FedOCR-Hash$_{1}$, it has fewer parameters than ASTER-FL, which benefits communication efficiency in federated learning. As shown in Fig.~\ref{fig:compression}, FedOCR-Hash$_{1}$ has a higher accuracy than ASTER-FL when little communication bytes are uploaded.

Fig.~\ref{fig:compression} illustrates the accuracy curves of different models on IIIT5k versus uploaded bytes in federated training procedures. FedOCR-Hash$_{\gamma}$ with a smaller compression ratio $\gamma$ achieves higher accuracy when limited communication bytes are uploaded, and it shows greater advantages in communication efficiency. The advantage of our FedOCR-Hash$_{\gamma}$ will be more distinctive when more local clients participate in our FedOCR.
Considering both Tab.~\ref{table:paramters} and~\ref{table:federated}, FedOCR-Hash$_{1/4}$ with an appropriate compression ratio $\gamma$ shows a significant overall performance in communication efficiency and accuracy of federated learning. Only $13.38$ megabytes are required to be transmitted by each participant, which results in a faster parameter transmission with the same communication bandwidth.

Benefited from lightweight models and hashing techniques, our federated scene text recognition framework shows a comparable performance and advantages in communication efficiency. Considering plenty of participants and the unstable data transmission network in the real world, our FedOCR has great potential in practical application deployment.

\section{Conclusion and Future Work}
In this paper, we reveal the problem of data privacy in scene text recognition and address the difficulty in utilizing decentralized datasets distributed on local devices with federated learning. To the best of our knowledge, we propose the first federated scene text recognition framework named FedOCR. In our FedOCR, we succeed in training a shared text recognizer collaboratively with decentralized datasets and avoid violating rules of data privacy. 
Benefited from lightweight models and hashing techniques, we reduce communication costs to a large extent and provide higher-level privacy-preserving against the honest-but-curious global server.
In terms of taking advantage of tremendous decentralized real-world data in practice, our communication-efficient federated learning framework for scene text recognition shows intriguing merits.

Recently, the domain shift in scene text recognition has attracted great interest in academia, and some methods are proposed, such as GA-DAN~\cite{zhan2019ga} and SSDAN~\cite{zhang2019sequence}. Notably, the domain shift occurs in federated learning for scene text recognition as well, which leads to a deterioration on the global accuracy.
Hence, we are working on the domain adaptation of decentralized datasets within our framework.

{
\bibliography{aaai22}
}
\end{document}